%% file: main.tex
\documentclass{article}

\usepackage[square,numbers]{natbib}

\usepackage[preprint]{neurips_2022}

\usepackage[utf8]{inputenc} 
\usepackage[T1]{fontenc}    
\usepackage{hyperref}       
\usepackage{url}            
\usepackage{booktabs}       
\usepackage{amsfonts}       
\usepackage{nicefrac}       
\usepackage{microtype}      
\usepackage{xcolor}         

\usepackage{amsmath}
\usepackage{amssymb}
\usepackage{mathtools}
\usepackage{amsthm}
\usepackage{multirow}
\usepackage{wrapfig}

\usepackage[ruled, vlined, linesnumbered, noend]{algorithm2e}

\DeclareMathOperator*{\argmin}{arg\,min}

\title{Delving into Effective Gradient Matching for Dataset Condensation}

\author{
  Zixuan Jiang, Jiaqi Gu, Mingjie Liu, David Z. Pan \\
  Department of Electrical and Computer Engineering \\
  The University of Texas at Austin \\
  Austin Texas, 78712 \\
  \texttt{\{zixuan, jqgu, jay\_liu\}@utexas.edu, dpan@ece.utexas.edu}
}

\begin{document}

\maketitle

\input{doc/0abstract}
\input{doc/1introduction}
\input{doc/2background}
\input{doc/3method}
\input{doc/4result}
\input{doc/5discussion}
\input{doc/6conclusion}



\bibliographystyle{abbrvnat}
\bibliography{reference}







\input{doc/appendix}
\end{document}

%% file: doc/0abstract.tex
\begin{abstract}
As deep learning models and datasets rapidly scale up, model training is extremely time-consuming and resource-costly.
Instead of training on the entire dataset, learning with a small synthetic dataset becomes an efficient solution.
Extensive research has been explored in the direction of dataset condensation, among which gradient matching achieves state-of-the-art performance.
The gradient matching method directly targets the training dynamics by matching the gradient when training on the original and synthetic datasets.
However, there are limited deep investigations into the principle and effectiveness of this method.
In this work, we delve into the gradient matching method from a comprehensive perspective and answer the critical questions of what, how, and where to match.
We propose to match the multi-level gradients to involve both intra-class and inter-class gradient information.
We demonstrate that the distance function should focus on the angle, considering the magnitude simultaneously to delay the overfitting.
An overfitting-aware adaptive learning step strategy is also proposed to trim unnecessary optimization steps for algorithmic efficiency improvement.
Ablation and comparison experiments demonstrate that our proposed methodology shows superior accuracy, efficiency, and generalization compared to prior work.
\end{abstract}

%% file: doc/1introduction.tex
\section{Introduction}
\label{section:introduction}

Large datasets are critical for the success of deep learning at the cost of computation and memory.
The high cost is unbearable when we train deep learning models with limited training time or memory budget.
For example, quick training is needed when training is a subtask.
When we perform neural architecture search~\cite{nas-survey}, hyper-parameter optimization~\cite{feurer2019hyperparameter, bergstra2013making}, or training algorithm design and validation,
we expect to obtain training performance quickly.
Another example is the training with limited storage space.
To overcome catastrophic forgetting in continual learning, we usually save partial samples for future training~\cite{overcoming-forgetting}.
There is a harsh constraint on the memory space when training on edge devices~\cite{edge-ml}.
Above all, it is a critical problem to achieve data efficiency in deep learning training.

As a traditional method to reduce the size of the training dataset, coreset construction defines a criterion for representativeness~\cite{coreset1, icarl, castro2018end, aljundi2019gradient, sener2017active} and then selects samples based on the criterion.
Coreset construction is used in many efficient and quick training tasks, e.g., accelerating hyperparameter search~\cite{shleifer2019using}, continual learning~\cite{NEURIPS2020_aa2a7737}.
Unlike the coreset construction method, dataset synthesis generates a small dataset, which is directly optimized for the downstream task.
Since it does not rely on representative samples, the dataset synthesis outperforms the coreset construction in the corresponding downstream task.

\citet{datasetdistillation} formulate the network parameters as a function of the synthetic training set and formulate the dataset condensation task as a bi-level optimization problem.
Specifically, the ultimate target is to train deep learning models on the synthetic training set from scratch such that the trained model can generalize to the original training dataset.
The authors minimize the training loss on the original large training data by optimizing the synthetic data.
Based on the formulation of the bi-level optimization problem, \citet{sucholutsky2019soft} extend the method by distilling both input and their soft labels.
\citet{such2020generative} propose to learn a generative teaching network, which generates synthetic data for training student networks.
\citet{nguyen2020dataset} use kernel ridge-regression to compress training datasets, enhancing the dataset distillation method.

\citet{datasetcondensation} propose to match gradients w.r.t. parameters when training examples come from synthetic and original datasets, respectively, to solve the bi-level optimization problem.
This method mimics the first-order loss landscape when the real training set is used and intuitively maximizes the landscape similarity via gradient matching.
By directly targeting the training dynamics, this optimization-aware methodology achieves the current state-of-the-art performance on dataset condensation.
However, the previous method does not deeply investigate the working principle in gradient matching, and the current matching flow has limited effectiveness and learning efficiency.

In this paper, we analyze the gradient matching method from a comprehensive perspective, including what, how, and where to match.
We enhance the gradient matching algorithm with three essential techniques to achieve higher efficiency and better task-specific performance.
We highlight our contributions as follows.
\begin{itemize}
    \item \textit{Multi-level matching.}
    We jointly explore intra-class and inter-class gradient matching to improve performance without extra gradient computation.
    \item \textit{Overfitting delaying.} 
    We propose to adopt a new type of gradient matching function to mitigate the overfitting issue on the synthetic training set to facilitate the optimization.
    We concentrate on the angle between the gradients, considering the magnitude simultaneously.
    \item \textit{Adaptive learning.}
    We update synthetic data against the parameter where overfitting happens.
    Thus, we can achieve the same performance with fewer parameter updates.
\end{itemize}

%% file: doc/2background.tex
\section{Background}

    

In this section, we first introduce the background of dataset condensation.
Then we describe the working principle of the gradient matching method as we will analyze and extend it in Section~\ref{section:method}.
Similar to the previous work, we take the classification task with balanced class distribution as an example.
The algorithm can be easily extended to other problems.

\textbf{Dataset condensation.} Dataset condensation is a task to generate a small synthetic training dataset $\mathcal{S}$ to mimic the model optimization behavior with the original training set $\mathcal{T}$.
Specifically, the network parameter $\theta$ is formulated as a function of the synthetic training set $\mathcal{S}$~\cite{datasetdistillation}.
\begin{equation}
    \label{equation:theta-S}
    \theta(\mathcal{S}) = \argmin_\theta \mathcal{L}(\mathcal{S}, \theta)
\end{equation}
$\mathcal{L}(\mathcal{S}, \theta) = \frac{1}{|\mathcal{S}|} \sum_{(x, y) \in \mathcal{S}} \ell (f_\theta(x), y)$ is the loss with synthetic dataset $\mathcal{S}$ and model parameter $\theta$.
$\ell$ is the task specific loss function, such as cross entropy loss in the classification task.
$f_\theta$ represents a deep learning model with parameter $\theta$.
Thus, the dataset synthesis problem can be written as the following bi-level optimization problem.
\begin{equation}
    \label{equation:bi-level}
    \min_\mathcal{S} \mathcal{L}\big(\mathcal{T}, \theta(\mathcal{S})\big) \quad \text{s.t.} \; \theta(\mathcal{S}) = \argmin_\theta \mathcal{L}(\mathcal{S}, \theta)
\end{equation}
We train a deep learning model $f$ using the synthetic training set $\mathcal{S}$ from scratch and obtain the optimal parameter $\theta(\mathcal{S})$.
The objective is to minimize $\mathcal{L}\big(\mathcal{T}, \theta(\mathcal{S})\big)$, the loss on the original large training set $\mathcal{T}$.
In other words, the original dataset $\mathcal{T}$ is the test dataset to verify the model $f_{\theta(\mathcal{S})}$.

\textbf{Gradient matching algorithm.}
Among all prior work on dataset condensation, the state-of-the-art performance has been achieved by gradient matching~\cite{datasetcondensation},
which directly encourages the training dynamics on the synthetic set to mimic that on the real training set.
The distance between gradients $\nabla_{\theta_t} \mathcal{L}(\mathcal{S}_t, \theta_t)$ and $\nabla_{\theta_t} \mathcal{L}(\mathcal{T}, \theta_t)$ is minimized, where $t=0,...,T$ is the time step.
If the gradients match, the training trajectories will be the same using gradient-based optimization methods.

\begin{figure*}
\centering
\includegraphics[width=0.9\textwidth]{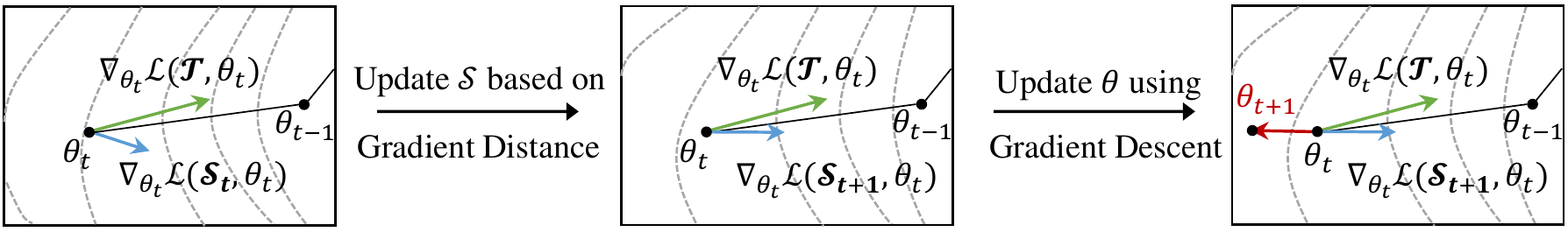}
\caption{One simplified inner loop iteration of dataset condensation with gradient matching algorithm.
We first update $\mathcal{S}$ to minimize the gradient distance.
Then network parameter is updated to imitate the training process.}
\label{figure:one-iteration}
\end{figure*}

Algorithm~\ref{alg:imporved-algorithm} shows the original gradient matching method.
For each iteration of the outer loop, the model parameters $\theta_0$ are initialized following the distribution of $P_{\theta_0}$.
Lines 5 $\sim$ 13 correspond one inner loop iteration, which is visualized in Figure~\ref{figure:one-iteration}.
The mini-batches are sampled in the same class when calculating the distance between these two gradients $\nabla_{\theta_t} \mathcal{L}(\mathcal{S}_t, \theta_t)$ and $\nabla_{\theta_t} \mathcal{L}(\mathcal{T}, \theta_t)$.
Afterwards, the distances for $C$ classes are accumulated to update the synthetic set,
\begin{equation}
\label{equation:s-update}
    \mathcal{S}_{t+1} = \mathcal{S}_{t} - \eta_{\mathcal{S}} \nabla_{\mathcal{S}_{t}} D\big(\nabla_{\theta_t} \mathcal{L}(\mathcal{S}_t, \theta_t), \nabla_{\theta_t} \mathcal{L}(\mathcal{T}, \theta_t)\big)
\end{equation}
where $D(a, b)$ is a function to measure the distance between two tensors, $\eta_{\mathcal{S}}$ is the learning rate.

In Lines 10 $\sim$ 13, the parameter $\theta_t$ is updated for $\zeta_\theta$ times to imitate the training process.
The gradient $\nabla_{\theta_t} \mathcal{L}(\mathcal{S}_{t+1}, \theta_t)$ instead of $\nabla_{\theta_t} \mathcal{L}(\mathcal{S}_{t}, \theta_t)$ or $\nabla_{\theta_t} \mathcal{L}(\mathcal{T}, \theta_t)$ is used to mimic the real training step to update the synthetic set.
The red arrow in Figure~\ref{figure:one-iteration} represents a gradient descent step.

The gradient matching method has several extensions.
\citet{extension-data-augmentation} extend it with differentiable data augmentation.
\citet{extension-composite} propose to learn a weighted combination of shared components to increase memory efficiency.
These extensions are orthogonal to our analysis and enhancements so that our method can be easily integrated into these extensions.

%% file: doc/3method.tex
\begin{algorithm*}[ht]
    \SetAlgoLined
    \SetKwInOut{Input}{Input}
    \SetKwInOut{Output}{Output}
    
    \Input{Original training set $\mathcal{T}$}
    Initialize $\mathcal{S}_0$ following Gaussian distribution\;
    \For(\tcp*[h]{outer loop: explore with different initialization}) {$k = 0, 1, ..., K-1$} { 
        Initialize model parameters $\theta_0 \sim P_{\theta_0}$\;
        \For(\tcp*[h]{inner loop}) {$t = 0, 1, ..., T-1$} { 
            \For{$c = 0, 1, ..., C-1$} {
                Sample mini-batches $B^{\mathcal{S}_t}_c \sim \mathcal{S}_t$, $B^\mathcal{T}_c \sim \mathcal{T}$ in the $c$-th class\;
                Compute gradients $g^{\mathcal{S}_t}_c =\nabla_{\theta_t} \mathcal{L}(B^{\mathcal{S}_t}_c, \theta_t), g^\mathcal{T}_c = \nabla_{\theta_t} \mathcal{L}(B^\mathcal{T}_c, \theta_t)$\;
            }
            $\texttt{loss} = \sum_{c=0}^{C-1} \textcolor{red}{\mathcal{D}}(g^{\mathcal{S}_t}_c, g^\mathcal{T}_c) + \textcolor{blue}{\lambda \textcolor{red}{\mathcal{D}}\big(\frac{1}{C}\sum_{c=0}^{C-1} g^{\mathcal{S}_t}_c, \frac{1}{C}\sum_{c=0}^{C-1} g^\mathcal{T}_c \big)}$\; 
            $\mathcal{S}_{t+1} = \mathcal{S}_{t} - \eta_{\mathcal{S}} \nabla_{\mathcal{S}_{t}} \texttt{loss}$\;
            $\theta_t^0 = \theta_t$\;
            \For(\tcp*[h]{update $\theta$ with adaptive learning steps}) {$i = 0, 1, ..., \textcolor{blue}{\zeta_\theta(t)}-1$} {
                $\theta_{t}^{i+1} = \theta_{t}^i - \eta_{\theta} \nabla_{\theta_t^i} \mathcal{L}(\mathcal{S}_{t+1}, \theta_t^i)$\;
            }
            $\theta_{t+1} = \theta_{t}^{i + 1}$\;
        }
    }
    \Output{Synthetic training set $\mathcal{S}$}
    \caption{Gradient matching algorithm. Our proposed methods are highlighted with color.}
    \label{alg:imporved-algorithm}
\end{algorithm*}

\section{Method}
\label{section:method}

In this section, we present our analysis and describe our improvement on the original algorithm.
We answer the following questions.
\textit{What, how, and where do we match in this gradient matching algorithm?}

\subsection{What we match: multi-level gradient matching}
\label{section:multi-level-gradient}
The original algorithm matches gradients of the mini-batch that samples in the same class.
Specifically, we sample mini-batches $B^{\mathcal{S}}_c \sim \mathcal{S}$, $B^\mathcal{T}_c \sim \mathcal{T}$ in the $c$-th class and calculate the gradients 
$g^\mathcal{S}_c =\nabla_{\theta_t} \mathcal{L}(B^{\mathcal{S}}_c, \theta_t), g^\mathcal{T}_c = \nabla_{\theta_t} \mathcal{L}(B^\mathcal{T}_c, \theta_t)$, respectively.
We minimize the distance between these intra-class gradients with accumulation.
\begin{equation}
    \texttt{loss}_\texttt{intra} = \sum_{c=0}^{C-1} D(g^\mathcal{S}_c, g^\mathcal{T}_c)
\end{equation}
Therefore, only the intra-class gradients are matched by using these intra-class mini-batches, missing the inter-class gradient information.
However, when we use either $\mathcal{S}$ or $\mathcal{T}$ to train the model, we usually use mini-batches that sample across different classes.
To mimic the realistic training process, we also need to match the gradients of these inter-class mini-batches.
We propose to match the gradients of these inter-class mini-batches in the following efficient way.

Since $\{g^\mathcal{S}_c\}_{c=0}^{C-1}$ has already been computed when calculating the intra-gradient distance, 
we can directly use them to compute the gradients for the mini-batches $\bigcup_{c=0}^{C-1} B^{\mathcal{S}}_c$ as follows.
\begin{align}
    \label{equation:gradients-union}
    g^\mathcal{S}_\cup = \nabla_{\theta_t} \mathcal{L}(\bigcup_{c=0}^{C-1} B^{\mathcal{S}}_c, \theta_t) = \frac{\sum_{c=0}^{C-1} |B^{\mathcal{S}}_c| g^\mathcal{S}_c}{\sum_{c=0}^{C-1} |B^{\mathcal{S}}_c|}
\end{align}
If the mini-batch $B^{\mathcal{S}}_c$ shares the same size, we can further simplify Equation~\eqref{equation:gradients-union} and obtain $g^\mathcal{S}_\cup = \frac{1}{C}\sum_{c=0}^{C-1} g^\mathcal{S}_c$.
We can also assign different weights to different mini-batches $B^{\mathcal{S}}_c$ to mimic the original training set $\mathcal{T}$ if the class distribution is not balanced in $\mathcal{T}$.
$g^\mathcal{T}_\cup = \nabla_{\theta_t} \mathcal{L}(\bigcup_{c=0}^{C-1} B^{\mathcal{T}}_c, \theta_t)$ can be computed in the same way.
In this way, we do not perform extra forward and backward computations to calculate gradients for the inter-class mini-batches $\bigcup_{c=0}^{C-1} B^{\mathcal{S}}_c$ and $\bigcup_{c=0}^{C-1} B^{\mathcal{T}}_c$.

With these inter-class gradients, we add a new term in the gradient matching loss as shown in Equation~\eqref{eq:multi-level}.
\begin{equation}
    \label{eq:multi-level}
    \underbrace{\sum_{c=0}^{C-1} D(g^{\mathcal{S}_t}_c, g^\mathcal{T}_c)}_{\texttt{loss}_\texttt{intra}} + \lambda \underbrace{D\big(\frac{1}{C}\sum_{c=0}^{C-1} g^{\mathcal{S}_t}_c, \frac{1}{C}\sum_{c=0}^{C-1} g^\mathcal{T}_c \big)}_{\texttt{loss}_\texttt{inter}}
\end{equation}
The first term is the intra-class gradient matching loss $\texttt{loss}_\texttt{intra}$, which is used in the original method.
We add a new term of inter-class gradient matching loss, with $\lambda$ being the weight to balance these two terms.
In this multi-level gradient matching loss, we consider both the intra-class and inter-class information.
Through experiments, we find that the multi-level gradient matching has better performance than either the intra-class or inter-class counterpart.
Experimental results are shown in Section~\ref{section:result-mini-batch}.

\subsection{How we match: angle and magnitude}

In the original gradient matching algorithm~\cite{datasetcondensation},
the authors propose to decompose the matching loss layer by layer
\begin{equation}
\small
\mathcal{D}\big(\nabla_{\theta} \mathcal{L}(\mathcal{S}, \theta), \nabla_{\theta} \mathcal{L}(\mathcal{T}, \theta)\big)  = \sum_{l=1}^L d\big(\nabla_{\theta^{l}} \mathcal{L}(\mathcal{S}, \theta), \nabla_{\theta^{l}} \mathcal{L}(\mathcal{T}, \theta)\big)
\end{equation}
where $L$ is the number of layers. 
For each layer, negative cosine similarity is used as the distance between two tensors,
\begin{equation}
d(A, B) = \sum_{i=1}^{\texttt{out}} \big(1 - \frac{A_i \cdot B_i}{\lVert A_i \rVert \lVert B_i \rVert}\big),
\end{equation}
where \texttt{out} is the number of output channels.
For example, the weights and the corresponding gradients of a 2D convolution layer has the shape of (\texttt{out}, \texttt{in/groups}, \texttt{h}, \texttt{w})
~\footnote{\texttt{out} and \texttt{in} are the number of output channels and input channels.
\texttt{groups} is the number of blocked connections from input channels to output channels.
\texttt{h} and \texttt{w} are kernel height and width, respectively.}.
We reshape the gradients as (\texttt{out}, $\texttt{in/groups} \times \texttt{h} \times \texttt{w}$) and compute the cosine similarity for each output channels.
This distance function considers the layer-wise structure and output channels, enabling a single learning rate across all layers.
By maximizing the cosine similarity between gradients, the $\mathcal{S}$ is expected to lead the parameter in a correct direction.

\begin{wrapfigure}{r}{0.4\textwidth}
    \centering
    \includegraphics[width=0.38\textwidth]{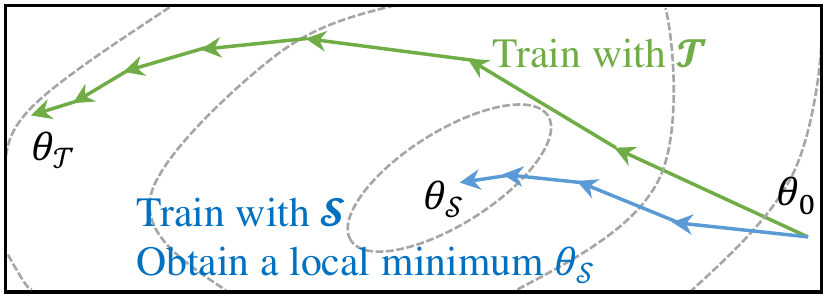}
    \caption{Optimization trajectories of training from scratch using $\mathcal{S}$ and $\mathcal{T}$.}
    \label{figure:train-from-scratch}
\end{wrapfigure}

However, in this distance matching loss, only the angle between gradients is considered, with the magnitude ignored.
This is a critical issue when we train a network $f$ from scratch for evaluation using the resultant $\mathcal{S}$.
Figure~\ref{figure:train-from-scratch} visualizes the training process using $\mathcal{S}$ and $\mathcal{T}$ respectively.
Since $|\mathcal{S}|$ is usually very small, it is a severe challenge that the deep learning model can easily remember the samples, which induces overfitting and a bad generalization.
The norm of gradient $\lVert \nabla_{\theta} \mathcal{L}(\mathcal{S}, \theta) \rVert$ degrades quickly during the training process.
In few gradient descent steps, we will be stuck in a local minimum, where $\nabla_{\theta} \mathcal{L}(\mathcal{S}, \theta) = 0$.

It is meaningless to match the angle when either of the gradient norm are small.
The right direction cannot help us escape the local minimum.
Thus, we have to consider the magnitude of the gradient vectors in this distance function.
For example, we can consider the Euclidean distance between two vectors $A_i$ and $B_i$,
\begin{equation}
\label{equation:cosine_euclidean}
    d(A, B) = \sum_{i=1}^{\texttt{out}} \big(1 - \frac{A_i \cdot B_i}{\lVert A_i \rVert \lVert B_i \rVert} + \lVert A_i - B_i \rVert \big)
\end{equation}
We also try other distance functions that consider the magnitude and show the results in Section~\ref{section:result-distance}.

\subsection{Where we match: adaptive learning steps}
In the $t$-th iteration of the inner loop, $\theta_t$ is used to update $\mathcal{S}_{t}$, and $\mathcal{S}_{t+1}$ is used to update $\theta_t$.
Here comes the question in this sequential update.
When we update $\mathcal{S}_{t}$, how many gradient descent steps do we perform such that $\mathcal{S}_{t+1}$ is a \textit{good} training set for $\theta_t$?
Similarly, how many gradient descent steps do we conduct when we update $\theta_t$ such that $\theta_{t+1}$ is a \textit{good} point for $\mathcal{S}_{t+1}$?

In the original algorithm, the authors provide their answers by setting the number of gradient descent steps $\zeta_{\mathcal{S}}$, $\zeta_\theta$ empirically.
Namely, we have the following update flows.
\begin{align}
    \mathcal{S}_t = \mathcal{S}_t^0 \rightarrow \mathcal{S}_t^1 \rightarrow \mathcal{S}_t^2 \rightarrow ... \rightarrow \mathcal{S}_t^{\zeta_{\mathcal{S}}} = \mathcal{S}_{t+1} \\
    \theta_t      = \theta_t^0      \rightarrow \theta_t^1      \rightarrow \theta_t^2      \rightarrow ... \rightarrow \theta_t^{\zeta_\theta}             = \theta_{t+1}
\end{align}

We present our understanding of these two hyper-parameters.
A change on $\mathcal{S}$ may induce a non-negligible update in the gradient $\nabla_{\theta} \mathcal{L}(\mathcal{S}, \theta)$, which is large enough for a update in the parameter.
Moreover, $\mathcal{S}$ should not be updated many times at one parameter $\theta$ to avoid overfitting.
Hence, the original setting of $\zeta_{\mathcal{S}}=1$ is a good choice.

Updating $\theta_t$ is an imitation of the training process.
The synthetic dataset $\mathcal{S}$ is the training dataset, while the original training dataset $\mathcal{T}$ serves as the validation dataset.
In particular, after one update from $\theta_t^i$ to $\theta_t^{i+1}$, 
we usually have a smaller training loss $\mathcal{L}(\mathcal{S}_{t+1}, \theta_t^i) > \mathcal{L}(\mathcal{S}_{t+1}, \theta_t^{i+1})$.
However, it is unknown how the validation loss change. 
The relationship between $\mathcal{L}(\mathcal{T}, \theta_t^i)$ and $\mathcal{L}(\mathcal{T}, \theta_t^{i+1})$ is uncertain.
We can check if there exists overfitting after updating $\theta_t$.
The naive method of detecting overfitting is making comparison between $\mathcal{L}(\mathcal{T}, \theta_t^i)$ and $\mathcal{L}(\mathcal{T}, \theta_t^{i+1})$.
We can also check the gap between two loss terms $\mathcal{L}(\mathcal{T}, \theta_t^{i+1}) - \mathcal{L}(\mathcal{S}_{t+1}, \theta_t^{i+1})$ to decide whether if overfitting happens.

Ideally, we should update network parameters $\theta$ until overfitting happens.
If there is no overfitting, the $\mathcal{S}$ leads the parameter as $\mathcal{T}$ does, which means $\mathcal{S}$ is a good approximation of $\mathcal{T}$ in the perspective of gradient matching.
We do not need to update $\mathcal{S}$ in this case.
If overfitting happens, the $\mathcal{S}$ needs to be updated against the current parameter since the $\mathcal{S}$ has divergence from $\mathcal{T}$.
Hence, it is better to use dynamic and adaptive learning steps, which help us locate where we need to update $\mathcal{S}$ and improve the algorithm efficiency.

\begin{figure*}[htbp]
    \centering
    \includegraphics[width=0.73\textwidth]{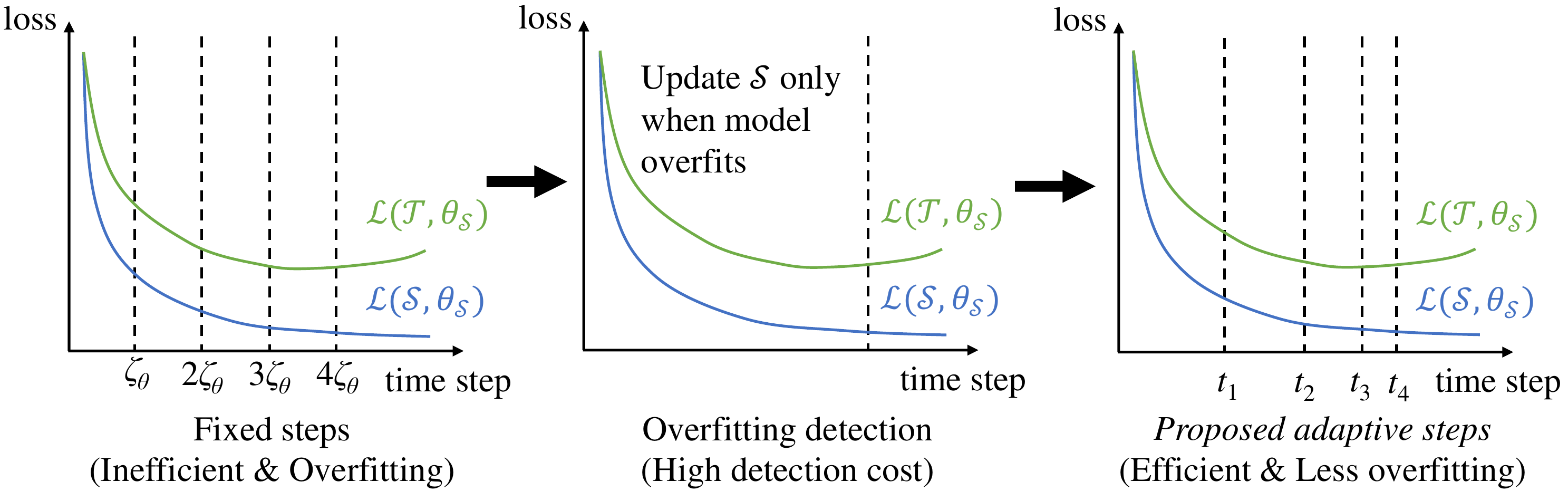}
    \caption{\textit{Left}. The original method updates $\mathcal{S}$ after updating network parameter $\theta$ for a fixed number of steps $\zeta_\theta$~\cite{datasetcondensation}. 
    \textit{Middle}. Ideally, $\mathcal{S}$ should be updated right after the model overfits.
    \textit{Right}. To avoid overhead on detecting overfitting, we propose to use adaptive learning steps $\zeta_\theta(t)$.}
    \label{figure:adaptive-learning-step}
\end{figure*}

Nevertheless, there is an overhead to detect the overfitting in real implementation.
For instance, we have to compute the loss term $\mathcal{L}(\mathcal{T}, \theta_t^{i+1})$ as the extra computation.
Therefore, we propose to run preliminary experiments and find when overfitting usually happens.
With the preliminary results, we make a schedule for $\zeta_\theta$.
In other words, let $\zeta_\theta$ be a function of the index of the current inner loop $t$ and we define this function from preliminary results to avoid the extra computation on overfitting detection.
This $\zeta_\theta(t)$ could help us locate where overfitting happens approximately.
Figure~\ref{figure:adaptive-learning-step} shows this improvement.

\subsection{Improved gradient matching flow}
Algorithm~\ref{alg:imporved-algorithm} shows our improvement on the gradient matching algorithm, with changes highlighted.
We match the multi-level gradients to consider both intra-class and inter-class information without extra gradient computation, as shown in Line 8.
We apply the new distance function, which considers magnitude, to avoid small gradients, which delays the overfitting.
We use a dynamic number of steps in Line 11 to improve the algorithm efficiency.
Ideally, the $\theta$ should be updated until overfitting happens where $\mathcal{S}$ diverges from $\mathcal{T}$.
We use a schedule $\zeta_\theta(t)$ to help us approximate when overfitting occurs.

%% file: doc/4result.tex
\section{Experiments}
In this section, we show an ablation study on our proposed techniques to validate their superiority to other variants and compare the test accuracy with prior arts. 

\begin{table*}[]
\small
\centering
\begin{tabular}{cccccc}
\hline
\textbf{Dataset}              & \textbf{\#Image/Class} & \textbf{intra-class}    & \textbf{inter-class}    & \textbf{interleaved}    & \textbf{multi-level} \\ \hline
\multirow{2}{*}{MNIST}        & 1         & \textbf{91.7$\pm$0.5} & 88.8$\pm$0.7          & \textbf{91.7$\pm$0.4} & 90.9$\pm$0.5             \\
                              & 10        & 97.4$\pm$0.2          & 96.9$\pm$0.1          & 97.1$\pm$0.1          & \textbf{97.6$\pm$0.1}    \\ \hline
\multirow{2}{*}{FashionMNIST} & 1         & 70.5$\pm$0.6          & 70.2$\pm$0.7          & \textbf{70.6$\pm$0.6} & \textbf{70.6$\pm$0.7}    \\
                              & 10        & 82.3$\pm$0.4          & 82.4$\pm$0.3          & 83.1$\pm$0.3          & \textbf{84.4$\pm$0.3}    \\ \hline
\multirow{2}{*}{SVHN}         & 1         & 31.2$\pm$1.4          & 29.8$\pm$0.7          & 30.8$\pm$1.6          & \textbf{32.9$\pm$1.2}    \\
                              & 10        & \textbf{76.1$\pm$0.6} & 72.7$\pm$1.0          & 75.4$\pm$0.7          & 75.5$\pm$0.7             \\ \hline
\multirow{2}{*}{CIFAR-10}      & 1         & 28.3$\pm$0.5          & \textbf{29.7$\pm$0.7} & 28.6$\pm$0.7          & \textbf{29.7$\pm$0.7 }            \\
                              & 10        & 44.9$\pm$0.5          & 46.7$\pm$0.5          & 45.9$\pm$0.6          & \textbf{48.6$\pm$0.5}    \\ \hline
\end{tabular}
\caption{Test accuracy (\%) with matching different gradients. New distance function and adaptive learning steps are disabled.}
\label{table:multi-level-gradients}
\end{table*}

\subsection{Settings}
\label{section:settings}

We follow the same settings with the original work for all the experiments.
Specifically, we use the same network architectures, datasets, and hyperparameters.
The difference is the improvement highlighted in Algorithm~\ref{alg:imporved-algorithm}.
We use five image classification datasets, MNIST~\cite{mnist}, FashionMNIST~\cite{fashionmnist}, SVHN~\cite{svhn}, CIFAR-10~\cite{cifar10} and CIFAR-100.
These datasets have a balanced class distribution.
There are 100 classes in the CIFAR-100 dataset, while other datasets have 10 classes.

We refer to the original work~\cite{datasetcondensation} and implementation~\footnote{\href{https://github.com/VICO-UoE/DatasetCondensation}{Link to the implementation}} for more details regarding experimental settings.
The results of the baseline method are from the original paper.

There are two phases in every single experiment.
We first use our algorithm to obtain a small synthetic training set $\mathcal{S}$ with a \textit{source} model.
In the second phase, we train a \textit{target} model with $\mathcal{S}$ from scratch and test the trained model on the original testing dataset.
For every experiment, we generate 2 sets of synthetic images and train 50 target networks.
The average and standard deviation of the test accuracy over these 100 evaluations are reported.

\subsection{Different gradient matching methods}
\label{section:result-mini-batch}

Our first improvement is to match the multi-level gradients, which combines the intra-class gradient matching and inter-class gradient matching as discussed in Section~\ref{section:multi-level-gradient}.
To demonstrate the efficacy of our proposed method, we make comparisons on the following four settings:
(1) intra-class gradient matching,
(2) inter-class gradient matching,
(3) matching these two gradients in an interleaved way\footnote{We match intra-class gradients in one iteration, and match inter-class gradients in the next iteration.},
and (4) multi-level gradient matching.
We use the same hyperparameters in these experiments, with $\lambda$ being the number of classes $C$, disabling the new distance function and adaptive learning steps.

Table~\ref{table:multi-level-gradients} lists the results on four datasets.
In most cases, the multi-level gradient matching achieves the best results.
Focusing on either intra-class gradient matching or inter-class gradient matching misses the other information.
Compared with minimizing the two matching losses in an interleaved way, the multi-level gradient matching is much more stable.

\subsection{Different distance functions}
\label{section:result-distance}

\begin{wraptable}{r}{0.5\textwidth}
\small
\centering
\begin{tabular}{cc}
\hline
\textbf{Distance Function} & \textbf{Test Accuracy} \\ \hline
$d_1(a, b) = 1 - (a \cdot b) / (\lVert a \rVert \lVert b \rVert)$                 & 32.9$\pm$1.2               \\
$d_2(a, b) = \lVert a - b \rVert$                 & 23.6$\pm$2.1               \\
$d_3(a, b) = \lVert a - b \rVert^2$                 & 24.3$\pm$1.8               \\
$100 d_4(a, b) = 100 d_3(a, b) / len(a)$                 & 23.5$\pm$1.4               \\
$d_1 + d_2$               & \textbf{34.5$\pm$1.9}               \\
$d_1 + d_3$               & 34.1$\pm$2.0               \\
$d_1 + 100 d_4$               & 34.0$\pm$1.4               \\ \hline
\end{tabular}
\caption{Test accuracy (\%) with different distance functions.}
\label{table:different-distance}
\end{wraptable}

Our second improvement is to use a new distance function.
Instead of only focusing on the angle between gradients, we also match the magnitude.
Here, we make comparisons on the following distance functions, with multi-level gradients enabled.
(1) Negative cosine similarity $d_1(a, b) = 1 - (a \cdot b) / (\lVert a \rVert \lVert b \rVert)$, 
(2) Euclidean distance $d_2(a, b) = \lVert a - b \rVert$,
(3) sum of the squared error $d_3(a, b) = \lVert a - b \rVert^2$,
(4) mean squared error $d_4(a, b) = d_3(a, b) / len(a)$,
Note that the $d_1$ focuses on the angle only, while the other functions consider angle and magnitude simultaneously.

We take the SVHN dataset with 1 image per class as an example.
Table~\ref{table:different-distance} lists the test accuracy when using these distance functions to generate synthetic sets.
We directly assign the same weight to these distance functions except that we set the weight of 100 for $d_4$.

We find that when using only one of these distance functions, the test accuracy with $d_1$ is the highest.
Our explanation is that the angle of the gradient is much more important than the magnitude when (stochastic) gradient descent and its variants are used.
However, when we combine $d_1$ and other magnitude-related distance functions, we get improvement compared with pure $d_1$.
Namely, we concentrate on the gradient direction while considering the magnitude to avoid being stuck in traps where the gradient norm is small.

\subsection{Adaptive learning steps}

\begin{table*}[ht]
\centering
\resizebox{0.99\textwidth}{!}{
\begin{tabular}{|c|c|ccc|cccc|c|}
\hline
\textbf{Dataset}       & \textbf{IPC} & \textbf{Random} & \textbf{Herding} & \textbf{DC Baseline} & \textbf{Ours-M} & \textbf{Ours-MD} & \textbf{Ours-MDO} & \textbf{Ours-MDA} & \textbf{Whole Training Set} \\ \hline\hline
\multirow{3}{*}{MNIST}          & 1   & 64.9$\pm$3.5    & 89.2$\pm$1.6     & 91.7$\pm$0.5   & 90.9$\pm$0.5   & \textbf{91.9$\pm$0.4}   & -              & -              & \multirow{3}{*}{99.6$\pm$0.0} \\
                                & 10  & 95.1$\pm$0.9    & 93.7$\pm$0.3     & 97.4$\pm$0.2   & 97.6$\pm$0.1   & \textbf{97.9$\pm$0.1}   & \textbf{97.9$\pm$0.1}   & \textbf{97.9$\pm$0.2}   &                               \\
                                & 50  & 97.9$\pm$0.2    & 94.8$\pm$0.2     & \textbf{98.8$\pm$0.2}   & 98.0$\pm$0.1   & 98.6$\pm$0.1   & 98.6$\pm$0.1   & 98.5$\pm$0.1   &                               \\ \hline
\multirow{3}{*}{FashionMNIST}   & 1   & 51.4$\pm$3.8    & 67.0$\pm$1.9     & 70.5$\pm$0.6   & 70.6$\pm$0.7   & \textbf{71.4$\pm$0.6}   & -              & -              & \multirow{3}{*}{93.5$\pm$0.1} \\
                                & 10  & 73.8$\pm$0.7    & 71.1$\pm$0.7     & 82.3$\pm$0.4   & 84.4$\pm$0.3   & \textbf{85.4$\pm$0.3}   & 84.6$\pm$0.3   & 84.2$\pm$0.3   &                               \\
                                & 50  & 82.5$\pm$0.7    & 71.9$\pm$0.8     & 83.6$\pm$0.4   & 87.8$\pm$0.2   & 87.4$\pm$0.2   & \textbf{87.9$\pm$0.2}   & \textbf{87.9$\pm$0.2}   &                               \\ \hline
\multirow{3}{*}{SVHN}           & 1   & 14.6$\pm$1.6    & 20.9$\pm$1.3     & 31.2$\pm$1.4   & 32.9$\pm$1.2   & \textbf{34.5$\pm$1.9}   & -              & -              & \multirow{3}{*}{95.4$\pm$0.1} \\
                                & 10  & 35.1$\pm$4.1    & 50.5$\pm$3.3     & 76.1$\pm$0.6   & 75.5$\pm$0.7   & 75.9$\pm$0.7   & \textbf{76.2$\pm$0.7}   & 75.9$\pm$0.7   &                               \\
                                & 50  & 70.9$\pm$0.9    & 72.6$\pm$0.8     & 82.3$\pm$0.3   & 82.2$\pm$0.2   & 82.9$\pm$0.2   & \textbf{83.8$\pm$0.3}   & 83.2$\pm$0.3   &                               \\ \hline
\multirow{3}{*}{CIFAR-10}       & 1   & 14.4$\pm$2.0    & 21.5$\pm$1.2     & 28.3$\pm$0.5   & 29.5$\pm$0.7   & \textbf{30.0$\pm$0.6}   & -              & -              & \multirow{3}{*}{84.8$\pm$0.1} \\
                                & 10  & 26.0$\pm$1.2    & 31.6$\pm$0.7     & 44.9$\pm$0.5   & 48.6$\pm$0.5   & 49.5$\pm$0.5   & 49.9$\pm$0.6   & \textbf{50.2$\pm$0.6}   &                               \\
                                & 50  & 43.4$\pm$1.0    & 40.4$\pm$0.6     & 53.9$\pm$0.5   & 58.5$\pm$0.5   & 58.6$\pm$0.4   & \textbf{60.0$\pm$0.4}  & 58.3$\pm$0.5   &                               \\ \hline  
\multirow{2}{*}{CIFAR-100}      & 1   & 4.2$\pm$0.3     & 8.4$\pm$0.3      & \textbf{12.8$\pm$0.3}   & 12.4$\pm$0.3   & 12.7$\pm$0.4   & -              & -              & \multirow{2}{*}{56.2$\pm$0.3} \\
                                & 10  & 14.6$\pm$0.5    & 17.3$\pm$0.3     & 25.2$\pm$0.3   & 30.8$\pm$0.3   & 28.0$\pm$0.4   & 29.5$\pm$0.3   & \textbf{31.1$\pm$0.3}   &                               \\ \hline
\end{tabular}
}
\caption{Ablation study in terms of the test accuracy (\%).
IPC is the number of image per class in $\mathcal{S}$.
\textit{Random} means that samples are randomly selected as the coreset.
\textit{Herding} selects samples whose center is close to the distribution center.
\textit{DC baseline} refers to the original work on dataset condensation.
$M$, $D$, $O$, $A$ represent multi-level gradient matching, new distance function, updating $\theta$ until overfitting, and adaptive steps, respectively.
$O$ and $A$ are not applicable when IPC is 1.}
\label{table:three-settings}
\end{table*}

Ideally, we should update $\mathcal{S}$ when it is no longer a good approximation of $\mathcal{T}$ in terms of gradient matching.
Thus, a criterion to detect overfitting is needed.
We try the naive overfitting criterion $\mathcal{L}(\mathcal{T}, \theta_t^i) < \mathcal{L}(\mathcal{T}, \theta_t^{i + 1})$.
In other words, if the validation loss increases, we will stop the parameter update and proceed to update $\mathcal{S}$.
With this setting, we have improved the test accuracy from $44.9\%$ to $45.7\%$ for 10 images per class of CIFAR-10.
However, we notice that this improvement is at the cost of overfitting detection, which is nontrivial in real implementation.
Therefore, we define a pre-defined schedule $\zeta_\theta(t)$ by observing when overfitting happens in the CIFAR-10 experiment above.
\begin{equation}
\label{equation:zeta-theta}
    \zeta_\theta(t) =
    \begin{cases}
    50 - 10t, & t < 4 \\
    10,  & 4 \leq t < 10 \\
    5,   & t \geq 10
    \end{cases}
\end{equation}
The reason $\zeta_\theta(t)$ is non-increasing is that we may encounter overfitting issues more frequently as training proceeds.
Hence, we need to update $\mathcal{S}$ at shorter intervals.

With this schedule, we can first proceed to where overfitting happens and then stay in this area.
Another advantage of this schedule is that it reduces the number of model parameter updates.
In the baseline method, the authors set $T = 1, 10, 50, \zeta_\theta = 1, 50, 10$ for synthesizing $1, 10, 50$ images per class.
Taking $T = 10$ as an example, the original algorithm updates $\theta$ 450 times in one iteration of the outer loop, while we only update it 190 times.

\subsection{Comparison with prior work}
\label{section:comparison}

In Table~\ref{table:three-settings}, we perform an ablation study on our methods with four settings to demonstrate the effectiveness of our proposed enhancement.
We name the four settings as \textit{Ours-M}, \textit{Ours-MD}, \textit{Ours-MDO}, and \textit{Ours-MDA},
where $M$, $D$, $O$, $A$ stand for \textbf{m}ulti-level gradient matching, new \textbf{d}istance function, updating $\theta$ until \textbf{o}verfitting, and \textbf{a}daptive steps, respectively.
We also add two coreset selection methods for comparison.
\textit{Random} means that samples are randomly selected as the coreset.
\textit{Herding}~\cite{herding1, herding2} selects samples whose center is close to the distribution center.
For a fair comparison, we evaluate our method on the same ConvNet model~\cite{convnet} as used in the original work~\cite{datasetcondensation}.
Both the source network and the target network are the ConvNet model.

For our method, we use the settings mentioned above.
We match the multi-level gradients as shown in Equation~\eqref{eq:multi-level} with $\lambda = C$, the number of classes.
We use the distance in Equation~\eqref{equation:cosine_euclidean}, the overfitting criterion $\mathcal{L}(\mathcal{T}, \theta_t^i) < \mathcal{L}(\mathcal{T}, \theta_t^{i + 1})$, and the adaptive learning step in Equation~\eqref{equation:zeta-theta}.
\footnote{Equation~\eqref{equation:zeta-theta} is from the observation on a CIFAR-10 experiment and we generalize the setting to other benchmarks.}
We use the same settings for all the benchmarks without further tuning~\footnote{An exception is that we use $d = d_1 + 0.1d_2 = 1 - (a \cdot b) / (\lVert a \rVert \lVert b \rVert) + 0.1 \lVert a - b \rVert $ for 50 images per class with CIFAR-10.}.
It is expected that we can achieve better results with better hyperparameters tuned for each benchmark. 
For instance, we can tune the distance function and the learning steps $\zeta_\theta(t)$.

Since the algorithm only runs a single inner loop, i.e., $T=1$, when the condensed dataset contains one image per class, the method of adaptive step has no impact in this setting.
In terms of test accuracy, our proposed multi-level gradient matching and angle-magnitude distance function outperform the baseline gradient matching method~\cite{datasetcondensation} in most benchmarks.
Our proposed adaptive learning step technique is a good approximation of overfitting detector since the results of \textit{Ours-MDO} and \textit{Ours-MDA} are similar.
With \textit{Ours-MDA}, we cut down unnecessary steps in the later optimization stage, leading to higher learning efficiency while maintaining our advantages in test accuracy.

Regarding the algorithm efficiency, the usage of multi-level gradients and new distance functions introduces less than $1\%$ extra computation time.
The adaptive learning step can reduce the computation time by $25\% \sim 30\%$ for the experiments with 10 images per class.

Although training on condensed datasets has a performance degradation, we argue that condensed datasets are usually used for quick training and training with limited resources.
If final training performance is the only objective, we have to conduct training on the whole dataset.
For reference, the baseline method achieves $64\%, 67\%, 71\%, 77\%, 83\%$ relative accuracy (the ratio compared to training on full dataset) with 50, 100, 200, 500, 1000 images per class on CIFAR-10 dataset.
We achieve $71\%, 74\%, 78\%, 84\%, 89\%$ relative accuracy in these settings.

One of the limitations is that we do not conduct experiments on large datasets for two reasons.
First, the exploration is quite computationally intensive so that we cannot handle these experiments.
Second, most of the previous work focuses on these benchmarks and we directly follow the same settings.
Moreover, we conduct experiments on tiny ImageNet (200 classes, image size $64 \times 64$, 1 image per class) with the same experiment settings as CIFAR-10/100.
The test accuracy (\%) of the baseline is $4.65\pm0.20$, while ours is $4.93\pm0.23$.

%% file: doc/5discussion.tex
\section{Discussion}

In this section, we present our qualitative analysis of the gradient matching method and our improvement.
When training a neural network using gradient descent or its variants, there is a reachable region in the parameter space.
We explore and enlarge this space from the initialization point throughout the training process.
Finally, we will select the best parameters in this region.
The gradient matching algorithm matches the first-order loss landscape of this reachable region, ignoring the unachievable parameter space.

We hypothesize that the gradient matching algorithm is actually the following optimization problem.
\begin{align}
\label{equation:hypothesis}
    \max_\mathcal{S} \min_{\theta, \theta_0} \quad & \lVert \theta - \theta_0 \rVert   \\
    \text{s.t.}                              \quad & \lVert \nabla_\theta \mathcal{L}(\mathcal{S}, \theta) \rVert < \epsilon \label{eq:local_minmum} \\
                                                   & \theta_0 \sim P_{\theta_0} \label{eq:init}
\end{align}
Here $\epsilon$ is a threshold such that the parameter satisfying $\lVert \nabla_\theta \mathcal{L}(\mathcal{S}, \theta) \rVert < \epsilon$ cannot escape this point with gradient descent method.
We maximize the distance between any pairs of the local minimum (Equation~\ref{eq:local_minmum}) and the initialization point (Equation~\ref{eq:init}).
Using gradient matching, we attempt to eliminate these reachable local minimums by updating the synthetic set $\mathcal{S}$.
Thus, we can proceed further and explore this reachable region when training with $\mathcal{S}$.
Each iteration of the outer loop of the Algorithm~\ref{alg:imporved-algorithm} is a depth-first search of this region.
We update $\mathcal{S}$ to escape from the traps in this region and update $\theta$ to explore and enlarge the reachable region.
We try different initialization to mimic the minimization in Equation~\eqref{equation:hypothesis}.

Our three improvement can be interpreted with this hypothesis.
The multi-level gradients consider both intra-class and inter-class information.
Matching inter-class gradients can help us synthesize images concentrating on the reachable region since we use inter-class mini-batches when using the resultant $\mathcal{S}$.
The intra-class gradient matching can accelerate the convergence.
With the distance function considering the magnitude, we can mitigate the internal local minimum and enlarge the reachable area.
The adaptive learning step can help us proceed to the internal local minimum or the boundary of this region to do updates efficiently.
Above all, our proposed techniques are used to remove the local minimum in the reachable region of the parameter space.

%% file: doc/6conclusion.tex
\section{Conclusion}

In this paper, we present our analysis of the gradient matching method for the dataset condensation problem.
Based on our analysis, we further extend the original algorithm.
We provide our answers to the question of \textit{what, how, and where we match in this gradient matching algorithm.}
We match the multi-level gradients to involve both intra-class and inter-class gradient information.
A new distance function is proposed to mitigate the overfitting issue.
We use adaptive learning steps to improve algorithm efficiency.
The effectiveness and efficiency of our proposed improvements are shown in the experiments.

For future directions, we may extend this algorithm into semi-supervised learning.
We would like to explore what labeled samples are needed in semi-supervised learning.
A general distance function and adaptive learning steps are also worth further investigation.

%% file: doc/appendix.tex
\appendix

\section{Extra discussions on experiments}

\subsection{Settings}
\label{appendix:settings}
We run all our experiments on a server with Intel Core i9-7900X CPU at $3.30 GHz$ and two NVIDIA Titan Xp GPUs (the CUDA version is 11.1.).

\subsection{Cross-architecture generalization}
\begin{table*}[h]
\small
\centering
\begin{tabular}{ccccccc}
\hline
{source\textbackslash{}target} & MLP      & ConvNet  & LeNet    & AlexNet   & VGG      & ResNet   \\ \hline
MLP                                               & 67.3$\pm$1.1 & 71.2$\pm$1.6 & 70.5$\pm$8.5 & 45.1$\pm$12.4 & 46.9$\pm$4.2 & 83.2$\pm$2.1 \\
ConvNet                                           & 72.7$\pm$1.6 & 91.9$\pm$0.4 & 84.1$\pm$2.2 & 84.1$\pm$2.4  & 84.3$\pm$1.5 & 90.4$\pm$0.4 \\
LeNet                                             & 62.8$\pm$1.6 & 87.2$\pm$0.7 & 81.8$\pm$1.9 & 80.8$\pm$3.6  & 75.1$\pm$2.5 & 88.1$\pm$1.1 \\
AlexNet                                           & 61.3$\pm$1.6 & 87.6$\pm$0.8 & 81.4$\pm$2.5 & 81.2$\pm$3.3  & 77.7$\pm$2.1 & 87.9$\pm$1.0 \\
VGG                                               & 63.8$\pm$2.3 & 89.3$\pm$0.7 & 82.4$\pm$2.4 & 82.5$\pm$2.8  & 81.1$\pm$2.2 & 89.4$\pm$0.8 \\
ResNet                                            & 62.2$\pm$1.8 & 82.3$\pm$2.3 & 80.4$\pm$3.2 & 80.1$\pm$2.5  & 75.7$\pm$3.2 & 87.4$\pm$1.0 \\ \hline
\end{tabular}
\caption{Cross-generalization test accuracy (\%). 
We condense the training set on one source network and test the resultant synthetic set on other target networks.}
\label{table:cross-generalization}
\end{table*}

Following the same setting as the original work, we present our experiments on the cross-architecture generalization with \textit{Ours-MD}.
We use the network architecture of multi-layer perceptron (MLP), ConvNet~\cite{convnet}, LeNet~\cite{lenet}, AlexNet~\cite{alexnet}, VGG-11~\cite{vgg}, and ResNet-18~\cite{resnet}.

Table~\ref{table:cross-generalization} shows the results when the source and target networks are different for 1 image per class of the MNIST dataset condensation.
We obtain a similar result to the original work~\cite{datasetcondensation}.
The datasets learned from the convolutional neural architectures (ConvNet, LeNet, AlexNet, VGG, and ResNet) generalize to the other convolutional networks.
Since the hyperparameters are searched based on the ConvNet, all the target networks achieve the highest test accuracy when trained from the dataset learned from ConvNet.
Moreover, the learned dataset can be used to validate the neural architectures.
For instance, we find that whatever the source network is, ResNet achieves one of the highest test accuracies among all target networks.